\documentclass[runningheads,a4paper]{llncs}

\usepackage{amssymb}
\usepackage{graphicx}
\usepackage{pdfpages}
\usepackage{amsmath}
\usepackage{url}
\usepackage{wrapfig}
\usepackage{subfig}

\newcommand{\keywords}[1]{\par\addvspace\baselineskip
\noindent\keywordname\enspace\ignorespaces#1}

\begin{document}

\mainmatter  

\title{3D Consistent Biventricular Myocardial Segmentation Using Deep Learning for Mesh Generation}
\titlerunning{3D Consistent Biventricular Myocardial Segmentation Using Deep Learning}  
%
\author{Qiao Zheng\inst{1}, Herv\'{e} Delingette\inst{1}, Nicolas Duchateau\inst{2}, Nicholas Ayache\inst{1}}
%
\authorrunning{Qiao Zheng et al.} 
%
\tocauthor{Qiao Zheng, Herv\'{e} Delingette, Nicolas Duchateau, and Nicholas Ayache}
%
\institute{
Universit\'{e} C\^{o}te d'Azur, Inria, France
\and
CREATIS, CNRS UMR 5220, INSERM U1206, France
}

\maketitle              

\begin{abstract}
We present a novel automated method to segment the myocardium of both left and
right ventricles in MRI volumes. The segmentation is consistent in 3D across the slices such that it can be directly used for mesh generation. Two specific neural networks with multi-scale coarse-to-fine prediction structure are proposed to cope with the small training dataset and trained using an original loss function. The former segments a slice in the middle of the volume. Then the latter iteratively propagates the slice segmentations towards the base and the apex, in a spatially consistent way. We perform 5-fold cross-validation on the 15 cases from STACOM to validate the method. For training, we use real cases and their synthetic variants generated by combining motion simulation and image synthesis. Accurate and consistent testing results are obtained.

\end{abstract}
%


%
\section{Introduction}
While most research about myocardial segmentation focuses on either left ventricle (LV) \cite{ave:khe} or right ventricle (RV) \cite{ave:khe:2} segmentation on 2D slices,
there is a great need for 3D-consistent biventricular (BV) segmentation, in which LV and RV together are segmented. 3D-consistent BV segmentation provides not only consistency, but also robustness on poor-quality slices (e.g. near apex). Its output may then be used to generate complete meshes. These advantages are missing from most other methods. For example the model of \cite{tra} is not very capable of segmenting the slices near apex.
The lack of publicly available automatic BV mesh generation tool and the success of deep learning on medical image analysis \cite{zho:gre} motivate us to develop this method based on two neural networks
: the former segments a single slice in the volume and the latter propagates the segmentation to the other slices.

\begin{figure}[!]
\centering
\makebox[\textwidth][c]{\includegraphics[width=1.0\textwidth, height=0.2\textheight]{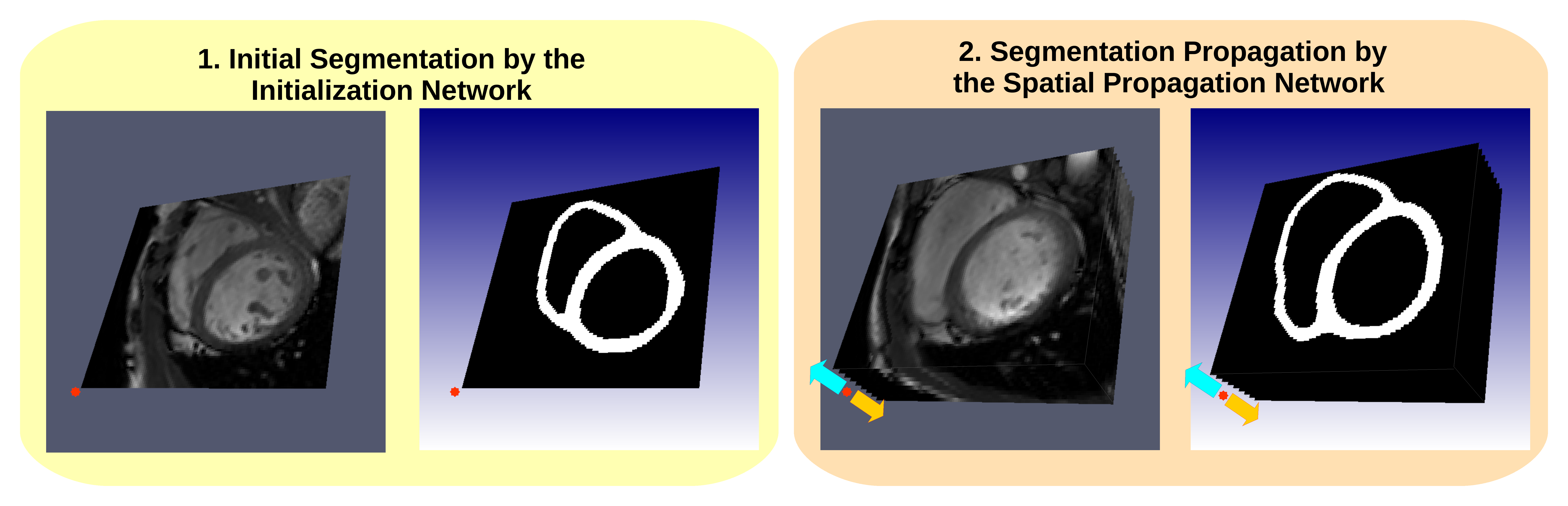}}
\centering
\caption{Overview of the proposed method. (1) Segmentation of a single slice at the middle of the volume by the initialization network. (2) Propagation of predicted mask towards base and apex by the spatial propagation network.}
\end{figure}

%
%
The most competitive BV myocardial segmentation methods are the automatic ones. 
Shape-constrained deformable models are applied on a dataset of 28 CT volumes in \cite{eca:pet}. The chain takes 22s for a volume. In \cite{zhe:bar} 4-chamber segmentation is performed using steerable features learned on a dataset of 457 annotated CT volumes. The speed is 4s per volume. The authors of \cite{wan:geo} apply marginal space learning and probabilistic boosting-tree on a dataset of 100 annotated MRI volumes to learn to jointly delineate LV and RV. It spends about 3s on each volume. We cannot compare directly with these methods on their reported error measures due to differences in datasets. In this paper we propose an effective pipeline based on 2 neural networks combining the assets of 2D (speed) and 3D (consistency). An original loss function is also applied for training. Our approach has the following advantages:
\begin{itemize}
\item Unlike the above-mentioned papers our networks are trained on a publicly available dataset STACOM \cite{tob:dec}. 
\item The dataset we use of 15 annotated volumes is much smaller than the above-mentioned datasets. Our method is data-efficient as its generalize even with small training sets.
\item Unlike the above methods, our method is model-free. Anyone familiar with deep learning may implement our networks without difficulty.
\item Our method takes about 3s to segment a volume. This is the same as the fastest one in the above methods.
\item Compared to MRI images, CT images usually have much better resolution and stronger heart/background contrast. Working with MRI images, we actually solve a more challenging version of segmentation problems. 
\end{itemize}
\section{Approach}
%
%
\subsubsection{Data preprocessing.}
%
The method takes short-axis-view MRI images as input. 
We process 3D stacks of 2D slices, cropped around the heart.
%
As standardization the cropped volume is resampled into an isotropic volume by linear interpolation.
Furthermore, 
we manually identify the base slice near mitral annulus. 
For images of reasonably good quality (e.g. STACOM) the segmentation can be initialized from any slice around the middle of the volume. So in this paper, for testing on the 3 testing cases of STACOM in each fold, the initialization slice is automatically chosen as the one in the middle of the volume. 

%

%
\subsubsection{Initial segmentation.}
We then apply the initialization network to segment the selected slice. The output is a mask of which each pixel is a probability (0 for background and 1 for myocardium).

\subsubsection{Spatial segmentation propagation.}
Then we apply the spatial propagation network to propagate segmentation masks. During upward (towards base) propagation, we suppose the slices up to that of index $z$ are already segmented. Taking this slice, its predicted mask and the next 4 slices ($z$+1 to $z$+4) as input, the spatial propagation network predicts the next 4 segmentation masks. The iteration stops at the base slice. Similarly, we downward (towards apex) propagate the segmentation masks. Thereby we complete the segmentation.

\begin{figure}[h]
\makebox[\textwidth][c]{%
    \begin{minipage}[t]{0.65\textwidth}
      \centering
      \includegraphics[width=\textwidth, height=0.5\textheight]{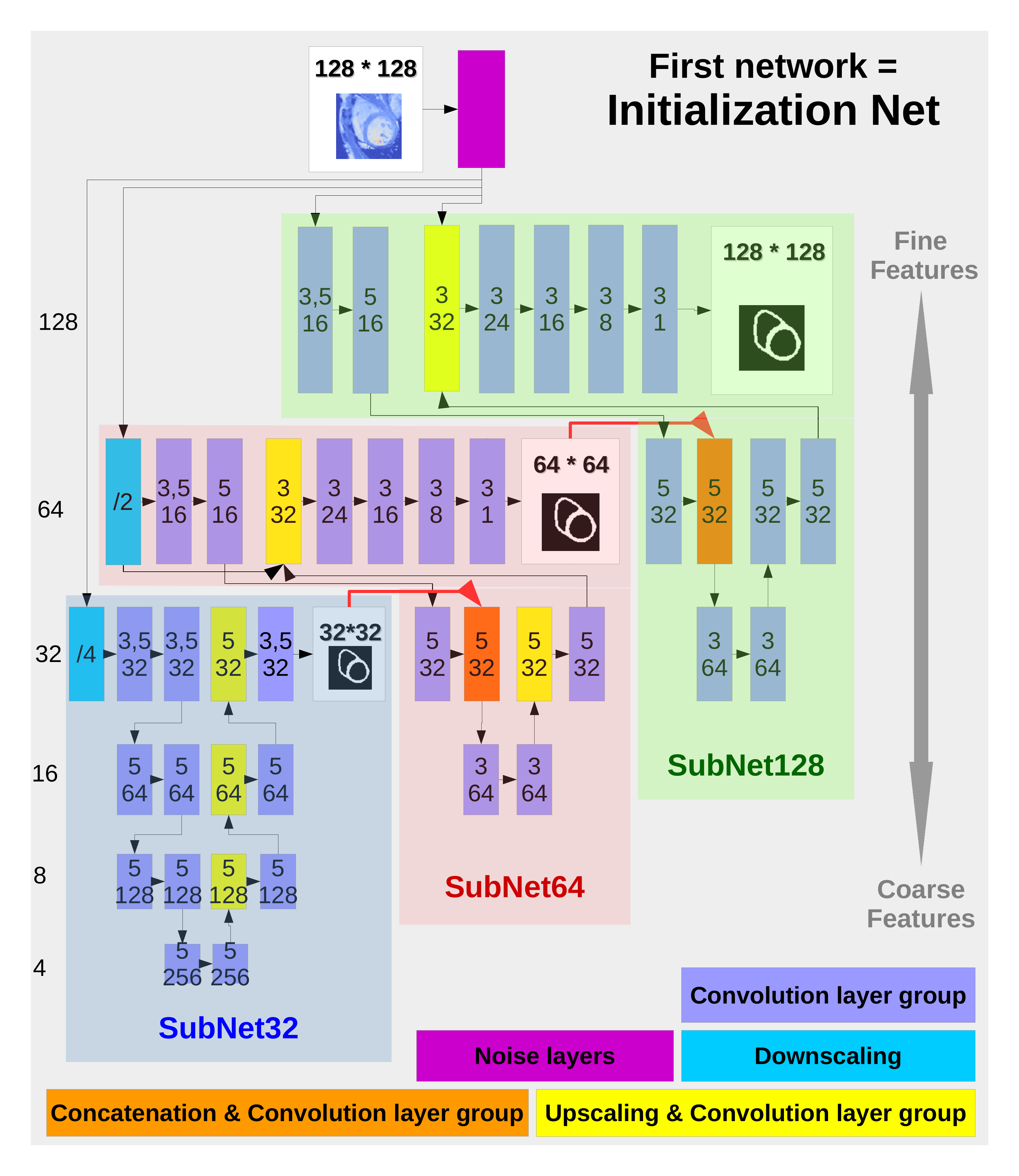}
    \end{minipage}%
    \hfill
    \begin{minipage}[t]{0.65\textwidth}
      \centering
      \includegraphics[width=\textwidth, height=0.5\textheight]{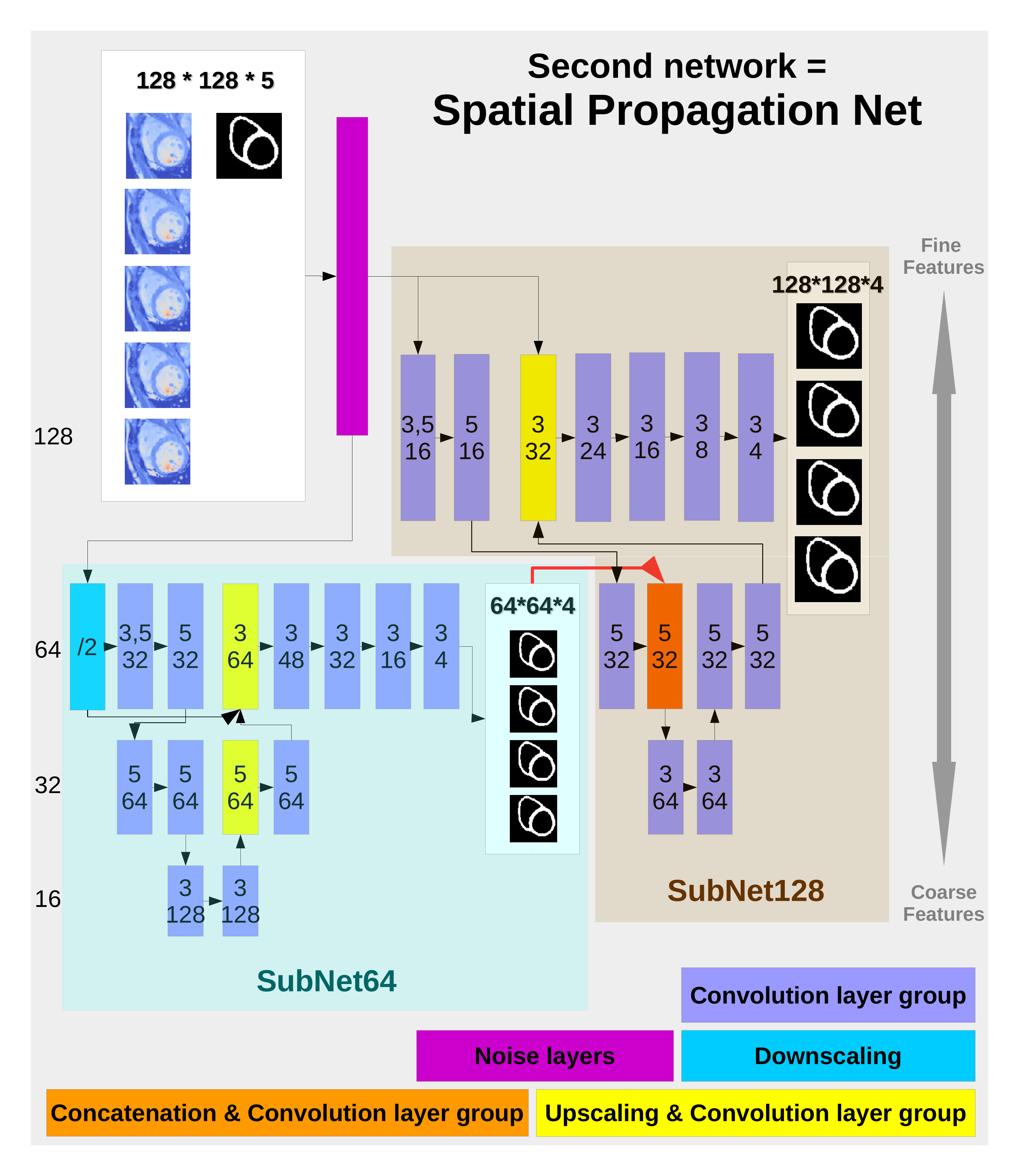}
      \label{fig:free-lunch}
    \end{minipage}%
}%
\caption{The two networks: the number marked to the left of each row is the size of output feature maps in the row; the upper number(s) in the rectangle of convolution layer is the filter size while the lower number indicates output channels; the number in rectangle of downscaling layer is the applied scale.}
\end{figure}

\section{Networks}
%
\subsubsection{Multi-scale coarse-to-fine prediction}
The two neural networks we use for this paper are characterized by multi-scale coarse-to-fine predictions. As presented in Fig. 2, the main body of the initialization network is separated into 3 sub-networks with input/output sizes 32, 64 and 128 respectively. SubNet32, taking a downsampled slice of size 32 as input, outputs a predicted mask of the same size. Then SubNet64 takes a downsampled slice of size 64 and incorporates the predicted mask of size 32 to make a prediction of size 64. Similarly, SubNet128 outputs the final predicted mask of size 128. 
During training, 3 loss functions which compare the outputs of SubNet32, SubNet64 and SubNet128 with the ground truth masks of size 32, 64 and 128 respectively are applied. The spatial propagation network, consisting of SubNet64 and SubNet128, has analogous structure and loss functions for training.

\subsubsection{Loss function.}
The networks are trained by stochastic gradient descent. An original loss function is designed to overcome numerical instability and class imbalance during training. We call it \textit{stabilized and class-balanced cross entropy loss}, where pixel-wise losses are added to work with a total loss. For each pixel, suppose the predicted probability is \textit{p} and the ground truth is \textit{g}. The pixel loss is
\begin{equation}
pixelLoss = \begin{cases}
$0$ &\text{  if $|g - p'|$ $<$ t,}\    \\
-log(p') &\text{  if $g = 1$ and $p'$ $\le$ $1 - t$,}\    \\
-log(1 - p') &\text{  if $g = 0$ and $p'$ $\ge$ $t$,}\    
\end{cases}
\end{equation}
with
\begin{equation}
p' = ap + b  
\end{equation}
and \textit{a}, \textit{b} and \textit{t} are parameters such that \textit{a} $>$ \textit{0}, \textit{b} $>$ \textit{0}, \textit{a} $+$ 2\textit{b} = 1 and 0 $<$ \textit{t} $<$ 1. To roughly preserve the predicted probability, \textit{a} is set close to 1, \textit{b} and \textit{t} are set near 0. In this paper we empirically pick \textit{a} = 0.999, \textit{b} = 0.0005 and \textit{t} = 0.02. 

The purpose of applying (2) is to avoid computation of logarithm on any value too close to 0 while roughly reserving the predicted probability. Without it, poorly predicted values of \textit{p} may result in extremely large loss and gradient values, which may harm numerical stability of training and even cause overflow. 

On the other hand, there is a strong imbalance between myocardial and background pixel. The latter represents around 80\% of the image. With common loss functions, the overall training effect of background pixels dominates the effect of the myocardial pixels. It may hinder the network performance in recognizing the myocardium. Setting the loss to 0 in (1) whenever the prediction is close enough to ground truth reduces the effect. When the predicted probabilities for background are close enough to 0, our loss function stops ``pulling'' them further to 0 and instead focuses on ``pushing'' the probabilities on myocardium to 1.

\subsubsection{Convolution layer group.} The two networks mainly consist of convolution layer groups. In each group, a convolution layer is followed by a batch normalization layer
and a leaky ReLU layer of negative part coefficient 0.25.

\subsubsection{Data augmentation inside network.}
The first layers in the two networks are noise layers for data augmentation during training to make the networks more robust. They are removed in testing. Data augmentation includes randomly rotating input slices together and adding Gaussian and pepper-and-salt noise.

%

%

%

\begin{figure}[h]
\makebox[\textwidth][c]{%
    \begin{minipage}[t]{0.8\textwidth}
      \centering
      \includegraphics[width=1.00\textwidth, height=0.22\textheight]{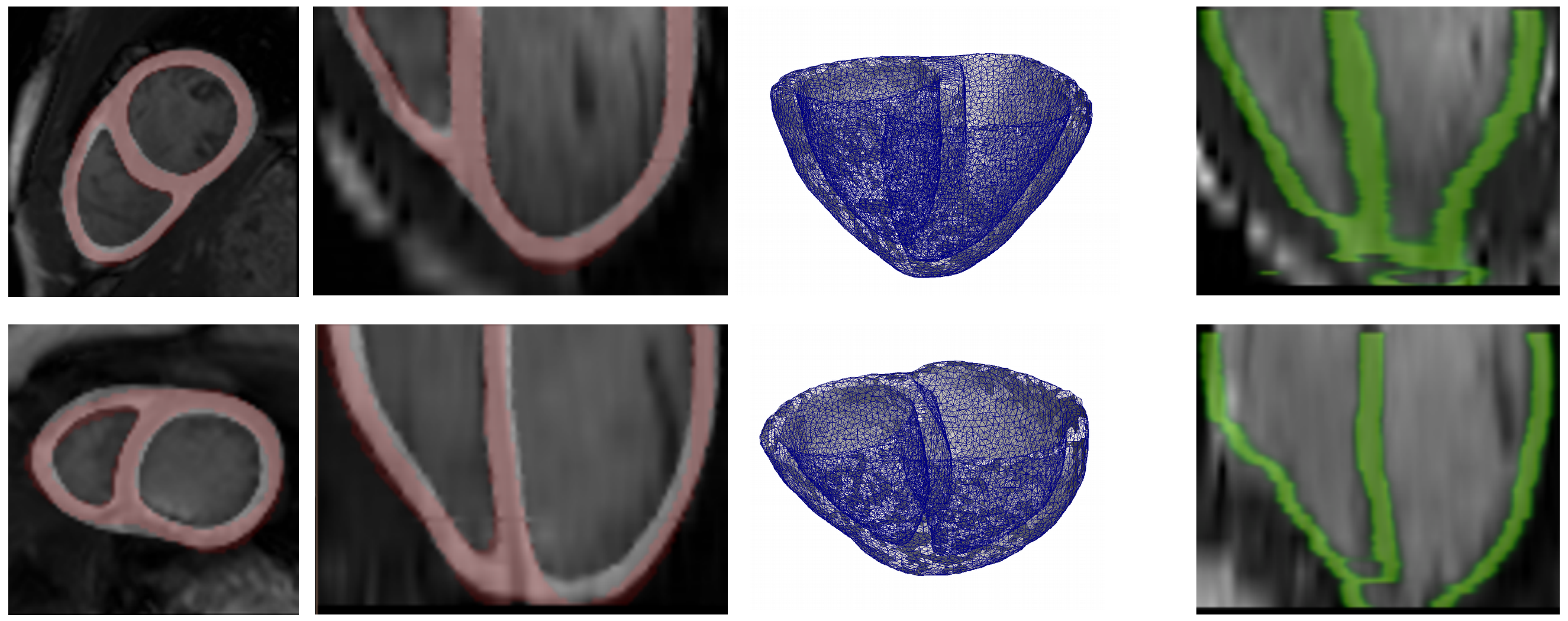}
    \end{minipage}%
    \hfill
    \begin{minipage}[t]{0.5\textwidth}
      \centering
      \includegraphics[width=0.95\textwidth, height=0.18\textheight]{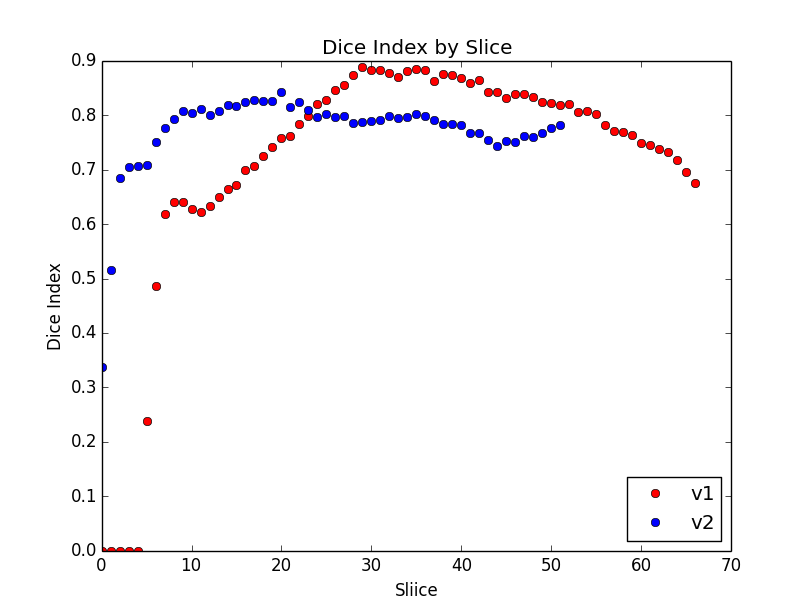}
      \label{fig:free-lunch}
    \end{minipage}%
}%
\caption{Left: prediction (white), ground truth (red) and generated meshes for the two tested cases from STACOM. Middle: segmentation results (green) without application of propagation. Right: slice-wise Dice Index on the test set.}
\end{figure}


%

%
\section{Experiments}

Our experiments involved an existing dataset with MRI image volume sequences: 15 subjects from STACOM \cite{tob:dec} (30 instants per cycle, with ground truth segmentation).
After resampling to isotropic volume of voxel size 1.25mm there are about 60 slices below the base in each volume. 
We divide the 15 cases into 5 groups of 3 cases. In each fold of the 5-fold cross-validation, the 3 cases of one picked group are used as testing. And the training set consists of the 12 cases from the remaining 4 groups.





\subsubsection{Training.}
As data augmentation to generate a large database with ground truth from a small database, we combine a motion simulation method and an image synthesis algorithm to generate realistic volume sequence variants of the training cases. Infarcted mesh motion sequences were simulated according to the scheme depicted in \cite{duc:dec}. Then the original volume sequences were warped to generate synthesized image variants using an algorithm inspired from \cite{pra:ser}. For each of the training subjects, 31 (1 healthy and 30 infarcted) 30-instant volume sequence variants were generated. In total 12*31*30=11160 volumes were used for training the spatial segmentation network in each fold of the 5-fold cross-validation. On the other hand, we observe better image/mesh coincidence around end-diastole than around end-systole in the synthesized sequences. Considering the trade-off between robustness (diversity in training set) and accuracy (image/mesh coincidence), we decide to train the initialization network only with volumes from 10 instants around end-diastole. Hence it is trained on 12*31*10=3720 volumes. Please note the augmentations of a same case remain similar. Our synthesized database is not comparable to real ones of similar size in terms of diversity and richness.

We use only the slices below the base in synthesized volumes to train the spatial propagation network. For the initialization net we also use these slices except the top 1/6 and the bottom 1/6 of them. The potentially poor image quality of slices near base and apex may cause additional inaccuracies.

The networks are trained by stochastic gradient descent with batch size 1 and learning rate 0.0001. The initialization network is trained for 300000 iterations. It takes about 23 hours in total on GPU. The spatial propagation network is trained for 600000 iterations, which together take roughly 44 hours.


%
\subsubsection{Testing.}
The method is tested on the end-diastole (the only instant where ground truth is available) of testing cases. It takes about 3s to segment a volume using GPU. The output probabilities are binarized to obtain myocardium/background segmentation taking 0.5 as threshold. We use the Dice index to measure performance. The 3D Dice indices (considering all pixels of all slices below base) are 0.7851 for v1 and 0.7817 for v2. The predicted masks and the ground-truth (axial and coronal views) as well as the BV meshes generated directly from the predicted masks using CGAL \footnote{http://www.cgal.org} are presented in the left part of Fig. 3. In the right part of Fig. 3, 2D Dice indices for both subjects change smoothly across slices, confirming the spatial consistency of our method. 

For comparison, if we use the initialization network to segment all the slices independently without propagation, the method not only loses the spatial consistency but also fails completely on the slices near apex (the middle part of Fig. 3). Our propagation method therefore appears crucial to maintain spatial consistency and reach accurate results even on the most difficult slices.

\section{Conclusion and Perspectives}
We demonstrate that our deep-learning-based automatic method for BV segmentation is robust, and combines the assets of 2D (speed) and 3D to provide spatially consistent meshes ready to be used for simulations. Besides, we proposed two original networks: (i)the initialization network predicts segmentation in a multi-scale coarse-to-fine manner; (ii)the second network propagates segmentation with spatial consistency. A novel loss function is also proposed to overcome class imbalance. For training, we use image synthesis as data augmentation. Meshes of high quality are generated. 
In the future, we will explore the capacity of neural networks in maintaining temporal consistency of segmentation. As we only use 15 subjects in this paper, significant improvement is expected if more data are added afterwards.


\bibliographystyle{splncs03}

%
%

\end{document}